\documentclass[11pt]{article}

\usepackage{acl}
\usepackage{float}

\usepackage{times}
\usepackage{latexsym}
\usepackage{amsmath}
\usepackage{rotating}
\usepackage[T1]{fontenc}
\usepackage[utf8]{inputenc}

\usepackage{microtype}

\usepackage{inconsolata}

\usepackage{graphicx}

\title{Maastricht University at AMIYA: Adapting LLMs for Dialectal Arabic using Fine-tuning and MBR Decoding}

\author{Abdulhai Alali \qquad {\bf Abderrahmane Issam} \\
        Department of Advanced Computing Sciences \\ 
        Maastricht University \\ 
  \small{\texttt{\{abdulhai.alali@student., abderrahmane.issam@\}maastrichtuniversity.nl}}}

\begin{document}
\maketitle
\begin{abstract}
% describe task -> approach -> results and findings
Large Language Models (LLMs) are becoming increasingly multilingual, supporting hundreds of languages, especially high resource ones. Unfortunately, Dialect variations are still underrepresented due to limited data and linguistic variation. In this work, we adapt a pre-trained LLM to improve dialectal performance. Specifically, we use Low Rank Adaptation (LoRA) fine-tuning on monolingual and English--Dialect parallel data, adapter merging and dialect-aware MBR decoding to improve dialectal fidelity generation and translation. Experiments on Syrian, Moroccan, and Saudi Arabic show that merging and MBR improve dialectal fidelity while preserving semantic accuracy. This combination provides a compact and effective framework for robust dialectal Arabic generation.

\end{abstract}

\section{Introduction}
% extend on task and it's challenges -> extend on approach -> extend on results, findings and limitations

Arabic dialects exhibit substantial variation in vocabulary, morphology, and syntax, making automated generation and translation challenging. Unlike Modern Standard Arabic (MSA), Dialectal Arabic (DA) is underrepresented in NLP resources, leading to difficulties in building models that produce fluent, semantically faithful, and dialectally authentic outputs \cite{alabdullah2025advancing}. The AMIYA Shared Task \cite{robinson-etal-2026-amiya} targets these challenges by evaluating LLMs on monolingual dialect generation and cross-lingual translation, emphasizing both dialect fidelity and instruction following.

To address these issues, we adapt a Large Language Model (LLM) using parameter-efficient fine-tuning on monolingual and English--Dialect parallel data. We train separate Low-Rank Adaptation (LoRA) adapters \cite{houlsby2019parameterefficienttransferlearningnlp, bapna-firat-2019-simple, hu2021lora} for each task (i.e. self supervised training on monolingual data and translation on parallel data), capturing dialectal surface forms and semantic grounding, and combine them using TIES-Merging \cite{yadav2023tiesmerging}. Additionally, we apply Minimum Bayes Risk (MBR) \cite{Bickel, kumar-byrne-2004-minimum, deguchi-etal-2024-mbrs} decoding with dialect-aware scoring to select outputs that maximize dialect authenticity during generation.

Our experiments show that merging monolingual and translation-based adapters improves the balance between dialectal fidelity and semantic accuracy. MBR decoding further enhances dialectal authenticity, leading to consistent gains over single-source fine-tuning and standard decoding. While our approach is effective, the following limitations persist: the training data is relatively small, dialect identification metrics may not capture subtle or informal usage, and MBR decoding increases inference time.

\section{AMIYA Shared Task}
% describes the task in more detail. citing al-qasida for challenges of generating dialectal output using lms. describes the metrics used for evaluation and evaluation data, the data used for training and evaluation

\subsection{Task Description}
The AMIYA Shared Task focuses on improving LLMs for Dialectal Arabic (DA), which remains significantly underrepresented compared to Modern Standard Arabic (MSA) \cite{bergman-diab-2022-towards}. Participants are asked to develop or adapt LLMs that can generate fluent, semantically faithful, and dialectally authentic Arabic across multiple regional varieties.

Systems are evaluated using the \textsc{AL-QASIDA} benchmark \cite{al-qasdia}, which measures dialectal fidelity, generation quality, and robustness to MSA--DA diglossia. Evaluation includes both monolingual dialect generation and cross-lingual settings, such as English--Dialect and MSA--Dialect translation. Performance is assessed using automatic metrics---primarily Arabic Dialect Identification And DIalectnes (ADI2) \cite{al-qasdia} for dialect fidelity and character-level F-score (chrF++) \cite{chrf, popovic-2015-chrf} for translation quality---as well as human judgments of fluency and instruction adherence.

\subsection{Datasets}

We participated in the closed data track of the shared task focusing on 3 out of 5 Arabic dialects provided by the task, namely: Syrian, Moroccan and Saudi Arabic. For each dialect, we combine two types of training data: Monolingual dialectal text which consists of unstructured sentences in the target dialect, and Machine Translation (MT) data with English and DA parallel text.

\begin{table}[h!]
\centering
\footnotesize
\setlength{\tabcolsep}{4pt}
\begin{tabular}{lll}
\hline
\textbf{Dialect (Type)} & \textbf{Dataset} & \textbf{\# Samples} \\
\hline
Syrian (Mono.) & Shami Corpus & 25{,}136 \\
Syrian (MT) & UFAL & 120{,}600 \\
Moroccan (Mono.) & DoDa & 10{,}000 \\
Moroccan (MT) & DoDa & 10{,}000 \\
Saudi (Mono.) & SDC & 14{,}891 \\
Saudi (MT) & SauDial & 1{,}000 \\
\hline
\end{tabular}
\caption{Datasets used per dialect and supervision type.
Shami Corpus~\cite{abu-kwaik-etal-2018-shami},
UFAL~\cite{krubinski-etal-2023-ufal},
DoDa~\cite{doda2023},
SDC~\cite{sdc2020},
SauDial~\cite{saudial2025}.}
\label{tab:dataset-overview-compact}
\end{table}

Table \ref{tab:dataset-overview-compact} summarizes the datasets and sample sizes used for fine-tuning across each dialect and supervision type. While more extensive data is available for most categories, we sub-sample the datasets to maintain computational efficiency and accelerate the experimental process.

% \subsection{Task Setup}

% The task evaluates the model’s ability to generate dialectally appropriate responses given input prompts. Prompts may be monolingual or cross-lingual, requiring the model to either produce fluent dialectal text directly or generate dialectal output conditioned on semantic content expressed in another language. This setup reflects real-world usage scenarios where dialectal generation must balance dialectal authenticity (i.e. correct vocabulary, morphology, and style), semantic faithfulness to the input, and robust instruction following.

\subsection{Evaluation Metrics}

Evaluation is performed using different metrics depending on the task setting:

\paragraph{Monolingual Dialect Evaluation.}
For monolingual generation, \textbf{ADI2} metric is used. ADI2 score was proposed in \citet{al-qasdia} to measure whether LLMs generate outputs that are dialectal, and whether they are faithful to the specific requested dialect. The level of dialectness is measured using Arabic Level of Dialectness of text (ALDI) \cite{aldi}, and the dialect class $C$ is predicted using a dialect identification baseline model from Nuanced Arabic Dialect Identification (NADI) 2024 shared task \cite{nadi}. More formally, ADI2 score \cite{al-qasdia}  is defined as:

\begin{equation}
\small
\text{score}_{\text{ADI2}}(y) = \text{score}_{\text{ALDi}}(y) * \text{score}_{\text{NADI}}(y)_C
\end{equation}

\paragraph{Cross-Lingual Evaluation.}
For translation-based and cross-lingual generation tasks, \textbf{chrF++} is used for evaluation. chrF++ is well suited for morphologically rich languages such as Arabic, where it captures fine-grained character overlap and is robust to spelling variation, making it appropriate for dialectal evaluation where orthographic inconsistency is common.

\section{System Description}
% describes finetuning model merging and mbr approaches and why we decided to use them

\subsection{LoRA Fine-tuning}
To incorporate dialectal knowledge into the base model, we use parameter-efficient fine-tuning with LoRA. Fine-tuning is performed separately for each dialect and task (i.e. self-supervised and translation), allowing the model to learn different types of information without intervention between them. Table \ref{tab:hyperparams} reports our training hyperparamters. These hyperparameters were chosen to ensure stable training under memory constraints while maintaining sufficient capacity for effective dialect adaptation.

\begin{table}[h!]
\centering
\small
\begin{tabular}{l c}
\hline
\textbf{Hyperparameter} & \textbf{Value} \\
\hline
Max. sequence length & 512 \\
Epochs & 5 \\
Learning rate & 3e-5 \\
Batch size (per device) & 2 \\
% Gradient Accumulation Steps & 4 \\
Effective batch size & 32 \\
Precision & BF16 \\
\hline
\end{tabular}
\caption{Key training hyperparameters.}
\label{tab:hyperparams}
\end{table}

\subsubsection{Monolingual Dialect Fine-tuning}

For monolingual adaptation, we fine-tune the model on raw dialectal text without any task-specific prompts. The data consists of standalone sentences written in each dialect, encouraging the model to naturally learn dialect-specific vocabulary, morphology, and sentence structure.

All sentences are tokenized using the JAIS \cite{jais2_2025} tokenizer with a fixed maximum sequence length. We train the model using a standard causal language modeling objective. To make fine-tuning efficient, we apply LoRA adapters \cite{bapna-firat-2019-simple, houlsby2019parameterefficienttransferlearningnlp, hu2021lora} to the attention layers of the model and update only these additional parameters during training. Furthermore, we rely on memory-optimization techniques such as gradient accumulation and gradient checkpointing, enabling larger effective batch sizes. The model is trained for multiple epochs using a standard optimization setup. This approach allows the model to adapt strongly to dialectal surface forms while preserving the general knowledge of the base model.

% , where the input tokens also serve as training targets. Padding is avoided to preserve sentence boundaries, and an end-of-sequence token is used when padding is required.

\subsubsection{Translation-Based Fine-tuning}

In addition to monolingual data, we fine-tune the model on an English–Dialect parallel dataset. This data exposes the model to aligned semantic content across languages, helping it associate dialectal expressions with their meanings and improving controllability during generation.

We frame translation as an instruction-following task in both directions: English$\rightarrow$Dialect and Dialect$\rightarrow$English. Each training example includes a natural language instruction specifying the translation direction, followed by the target output. During training, the loss is computed only on the output tokens, while the instruction tokens are masked. This encourages the model to follow instructions without learning to reproduce them.

Tokenization is performed with a fixed maximum sequence length. The same LoRA setup is used as in monolingual fine-tuning to ensure compatibility across training stages. Training is carried out with a more conservative optimization setup than monolingual fine-tuning, focusing on stable learning and semantic alignment rather than aggressive adaptation.

\subsubsection{Adapter Merging}

The monolingual and translation-based fine-tuning strategies provide complementary supervision. Monolingual fine-tuning emphasizes dialectal fluency and authenticity, while translation fine-tuning reinforces semantic faithfulness and cross-lingual grounding. By training separate LoRA adapters for each dataset, we preserve these distinct signals and later combine them using TIES-Merging. This separation enables fine-grained control over how different sources of supervision contribute to the final dialect-aware model and minimizes intervention between them.

\subsection{MBR Decoding with Dialect-Aware Scoring}
While fine-tuning and merging improve the model’s internal dialect representations, decoding decisions still play a crucial role in output quality. We therefore apply MBR decoding using the \textit{mbrs}\footnote{\url{https://github.com/naist-nlp/mbrs}} library to explicitly optimize for dialectness at inference time.

For each input prompt, the model generates a set of $N=20$ candidate responses via stochastic sampling, then each candidate is scored independently using the ADI2 metric. Finally, the candidate with the highest score is selected as the final output.

\subsection{Adapter Merging and MBR}
TIES-based adapter merging integrates complementary dataset supervision at the parameter level, producing a compact yet expressive dialect-aware model. MBR decoding complements this by enforcing dialectal fidelity at generation time, explicitly selecting outputs that maximize dialectness. Together, fine-tuning, TIES-Merging, and MBR decoding form a unified framework that yields more consistent and authentic dialectal generation than any single technique in isolation.

\section{Results}
% comparison between our final approach and baselines (with and without finetuning etc). also comparison between jais2 and llama...

The experiments were conducted exclusively on Syrian and Moroccan DA and subsequently applied to the remaining dialects for the final submission, which was trained separately per dialect. This section presents a detailed evaluation of our dialect-aware generation framework. We report results across model variants, data configurations, and decoding strategies, with the goal of understanding (1) the impact of model choice, (2) the role of different supervision signals, and (3) the effectiveness of adapter merging and MBR decoding. The evaluation datasets used are the default datasets provided by AL-QASIDA \footnote{\url{https://github.com/JHU-CLSP/al-qasida}}. 

\subsection{JAIS-2 vs. LLaMA 3.2}

We begin by comparing two LLMs, JAIS-2\footnote{\url{https://huggingface.co/inceptionai/Jais-2-8B-Chat}} \cite{jais2_2025} and LLaMA~3.2\footnote{\url{https://huggingface.co/meta-llama/Llama-3.2-3B-Instruct}} \cite{grattafiori2024llama3herdmodels}, to determine the most suitable backbone for Arabic dialect generation. For a fair comparison, both models are fine-tuned using the same data configuration: a merged setup that combines monolingual dialect data with English--Dialect parallel (MT) supervision. In addition, decoding is performed using MBR decoding with ADI2 score.

Models are evaluated using ADI2 for monolingual dialect generation and chrF++ for translation. Table~\ref{tab:jais_llama} presents the results. On monolingual dialect generation, LLaMA~3.2 achieves a substantially higher ADI2 score (i.e. 0.78), indicating strong dialectal surface realization and fluency. However, its performance drops sharply in translation, with a significantly low chrF++ score (i.e. 0.14), suggesting weak semantic alignment when translating from English into dialectal Arabic. In contrast, JAIS-2 exhibits a more balanced performance. While its ADI2 score (0.33) is considerably lower than that of LLaMA~3.2 for monolingual generation, JAIS-2 achieves a much higher chrF++ score (0.43) on MT-based generation. This indicates stronger semantic fidelity and better handling of translation supervision.
\begin{table}[ht]
\small
\centering
\begin{tabular}{lcc}
\hline
\textbf{Model} & \textbf{ADI2} & \textbf{chrF++} \\
\hline
LLaMA~3.2 & \textbf{0.78} & 0.14 \\
JAIS-2    & 0.33 & \textbf{0.43} \\
\hline
\end{tabular}
\caption{Comparison between JAIS-2 and LLaMA~3.2 after fine-tuning, TIES-Merging and generation with MBR decoding on \textbf{Syrian DA}}.
\label{tab:jais_llama}
\end{table}

On overall, although LLaMA~3.2 excels in dialectal surface form generation, its poor cross-lingual performance limits its usefulness for translation-driven dialect generation. Given our goal of building a dialect-aware system that remains reliable across both monolingual and cross-lingual scenarios, we select JAIS-2 as the backbone for all subsequent experiments.

\subsection{Effect of Fine-tuning Data and Adapter Merging}

In this section, we analyze the impact of different fine-tuning strategies on JAIS-2. We report the results of JAIS-2 base model, JAIS-2 fine-tuned on either monolingual or translation data, and JAIS-2 with TIES-Merging. The results in Table~\ref{tab:finetuning} show that monolingual fine-tuning substantially improves ADI2 scores, indicating stronger dialectal identity and linguistic conformity. Furthermore, fine-tuning on parallel translations significantly improves chrF++ scores, reflecting improved semantic faithfulness and cross-lingual grounding. More importantly, merging monolingual and MT models using TIES-Merging consistently improves the balance between dialectal authenticity and semantic accuracy, leading to the best chrF++ and the second best ADI2 score performance. 

% \begin{table}[h]
% \centering
% \small
% \begin{tabular}{lcc}
% \hline
% \textbf{Configuration} & \textbf{ADI2} & \textbf{chrF++} \\
% \hline
% JAIS-2 (Base) & 0.07 & 0.03 \\
% + Monolingual FT & 0.27 & 0.01 \\
% + MT FT & 0.16 & 0.09 \\
% + TIES-Merging & \textbf{0.38} & \textbf{0.20} \\
% \hline
% \end{tabular}
% \caption{Effect of monolingual and MT task fine-tuning and TIES-Merging on \textbf{(Syrian DA)}. Merging both datasets (i.e. TIES-Merging) leads to the best performance on both ADI2 and chrF++.}

% \label{tab:finetuning}
% \end{table}

\begin{table}[h]
\centering
\small
\begin{tabular}{lcc}
\hline
\textbf{Configuration} & \textbf{ADI2} & \textbf{chrF++} \\
\hline
JAIS-2 (Base) & 0.18 & 0.31 \\
+ Monolingual FT & \textbf{0.44} & 0.33 \\
+ MT FT & 0.26 & 0.42 \\
+ TIES-Merging & 0.38 & \textbf{0.44} \\
\hline
\end{tabular}
\caption{Effect of monolingual and MT task fine-tuning and TIES-Merging on \textbf{Moroccan DA}. Merging both tasks (i.e. TIES-Merging) leads to the best performance on chrF++ and the second best on ADI2 score.}
\label{tab:finetuning}
\end{table}

\subsection{MBR Decoding with Dialect-Aware Objectives}

While adapter merging improves the model’s internal representations, standard decoding does not always select the most dialectally appropriate output. To address this, we apply MBR decoding with different objectives. MBR requires a target metric to score the candidate generations. We experiment with using ADI2 to improve dialectal fidelity, chrF++ to improve cross-lingual grounding, and their combination. As shown in ~\ref{tab:mbr}, MBR decoding with ADI2 achieves the best overall balance, improving monolingual ADI2 to approximately 0.51 while also increasing MT ADI2 to 0.36. This represents a substantial improvement over standard decoding and demonstrates that dialect-aware reranking can recover dialectal authenticity without sacrificing semantic grounding. In contrast, chrF++-optimized MBR and the combined objective favor translation quality: they achieve higher chrF++ scores (0.42 and 0.41, respectively) but lead to a significant drop in monolingual ADI2 (0.24 and 0.29). These results indicate that chrF++-centric objectives bias the model toward more neutral or standardized Arabic forms, reducing dialectal distinctiveness. Based on these findings, and given our emphasis on dialect fidelity while maintaining acceptable translation performance, we select ADI2-based MBR decoding for the final submission.

\begin{table}[h]
\centering
\scriptsize
\begin{tabular}{lccc}
\hline
\textbf{Decoding Strategy} & \textbf{Mono ADI2} & \textbf{MT ADI2} & \textbf{chrF++} \\
\hline
Standard decoding & 0.38 & 0.27 & \textbf{0.44} \\
MBR (ADI2) & \textbf{0.51} & 0.36 & 0.40 \\
MBR (chrF++ ) & 0.24 & 0.30 & 0.42 \\
MBR (ADI2 + chrF++) & 0.29 & \textbf{0.37} & 0.41 \\
\hline
\end{tabular}
\caption{Effect of MBR decoding objectives on JAIS-2 after fine-tuning and TIES-MErging on \textbf{Moroccan DA}. Using ADI2 as an objective strikes the best balance in ADI2 and chrF++ performance.}
\label{tab:mbr}
\end{table}

\subsection{Final Submission}

Across all experiments, the best-performing configuration is: JAIS-2 independent fine-tuning on monolingual and translation data, merging with TIES technique, and decoding using MBR with ADI2 as an objective metric. This configuration achieves the strongest balance between dialectal authenticity and cross-lingual grounding, outperforming all alternatives in terms of combined monolingual and MT performance. Consequently, our \textit{primary} submission applies this methodology individually to the Moroccan, Syrian, and Saudi dialects. While joint training across all dialects may offer further gains, we defer this exploration to future work.

\section{AMIYA Shared Task Official Results}

Table~\ref{tab:auto_eval} summarizes the official automatic evaluation results of our \textit{primary} submission. The results demonstrate strong generalization across dialects and task settings, confirming that the combination of parameter-efficient fine-tuning, TIES-Merging, and dialect-aware MBR decoding provides a robust and effective solution for DA generation.

\begin{table}[h]
\centering
\scriptsize
\setlength{\tabcolsep}{4pt}
\begin{tabular}{lccccc}
\hline
\textbf{Dialect} & \textbf{ADI2} & \textbf{DA$\rightarrow$EN} & \textbf{EN$\rightarrow$DA} & \textbf{DA$\rightarrow$MSA} & \textbf{MSA$\rightarrow$DA} \\
\hline
Moroccan & \textbf{0.679} & 49.93 & 30.02 & 39.53 & 33.77 \\
Syrian   & 0.389 & \textbf{51.89} & \textbf{34.44} & \textbf{ 43.42 } & \textbf{40.33} \\
Saudi    & 0.464 & 0.03  & 19.82 & 37.21 & 24.23 \\
\hline
\end{tabular}
\caption{Automatic evaluation results using ADI2 and chrF++.}
\label{tab:auto_eval}
\end{table}
Besides automatic evaluation, our model generations were human evaluated for fluency and adherence to DA instructions. Table \ref{tab:human_eval} shows that our highest human evaluation performance is on the Moroccan dialect.
\begin{table}[h]
\centering
\small
\begin{tabular}{lcc}
\hline
\textbf{Dialect} & \textbf{Adequacy} & \textbf{Fluency} \\
\hline
Moroccan & \textbf{1.97}  & \textbf{3.37} \\
Syrian   & 1.146 & 2.625 \\
Saudi    & 1.122 & 2.378 \\
\hline
\end{tabular}
\caption{Human evaluation scores.}
\label{tab:human_eval}
\end{table}

Across submissions from other teams, our system achieved the highest ADI2 for Syrian and Saudi, and the highest chrF++ scores on translation from English and MSA into Syrian (ENG$\rightarrow$DA and MSA$\rightarrow$DA). On Moroccan Arabic, our model performs best on the human evaluation of fluency.

% MBR decoding optimized for ADI2 yields higher overall ADI2 scores, particularly for Syrian and Saudi dialects, even when DA$\rightarrow$EN chrF++ is lower. This is expected, as ADI2 prioritizes dialectal fidelity over surface n-gram overlap. Higher chrF++ scores in EN$\rightarrow$DA and MSA$\rightarrow$DA indicate improved dialect generation, while increased variability in English outputs leads to lower chrF++. Human evaluations, especially fluency for Moroccan Arabic, align with this trend.

\section{Conclusion}
We presented a method for improving Dialectal Arabic generation by combining fine-tuning on dialectal and translation data, LoRA adapter merging, and MBR decoding. This approach helps the model produce outputs that are both fluent in the target dialect and faithful to the input meaning. Our experiments across three dialects show that this combination works better than using any single technique on its own, providing a practical way to make LLMs more dialect-aware. 

\section*{Limitations}
This work has the following limitations: Our decoding strategy depends on automatic dialect identification (ADI2) scores, which may not always capture subtle or informal dialectal usage. The training data is limited in size and may not cover all linguistic variation within each dialect, especially code-switching and colloquial expressions. Finally, both ADI2 computation and minimum Bayes risk (MBR) decoding increase inference time. ADI2 requires an additional forward pass, and the cost of MBR grows with the number of candidate hypotheses generated per input, resulting in a several-fold slowdown compared to standard decoding. As a result, the approach may be less practical for real-time or low-latency applications.
% \section*{Acknowledgments}

% \bibliographystyle{acl_natbib}
\bibliography{ref}

\appendix

\end{document}